\documentclass[11pt]{article}

\usepackage[margin=1in]{geometry}
\usepackage{times}
\usepackage{graphicx}
\usepackage{booktabs}
\usepackage{amsmath}
\usepackage{amssymb}
\usepackage{hyperref}
\usepackage{url}

\title{mind\_call: A Dataset for Mental Health Function Calling with Large Language Models}

\author{
  Fozle Rabbi Shafi \\
  School of Computing, Queen's University \\
  Kingston, Canada \\
  \texttt{f.shafi@queensu.ca}
  \and
  M. Anwar Hossain \\
  School of Computing, Queen's University \\
  Kingston, Canada \\
  \texttt{ahossain@queensu.ca}
  \and
  Salimur Choudhury \\
  School of Computing, Queen's University \\
  Kingston, Canada \\
  \texttt{s.shoudhury@queensu.ca}
}

\date{} 

\begin{document}
\maketitle

\begin{abstract}
Large Language Model (LLM)–based systems increasingly rely on function calling to enable structured and controllable interaction with external data sources, yet existing datasets do not address mental health–oriented access to wearable sensor data. This paper presents a synthetic function-calling dataset designed for mental health assistance grounded in wearable health signals such as sleep, physical activity, cardiovascular measures, stress indicators, and metabolic data. The dataset maps diverse natural language queries to standardized API calls derived from a widely adopted health data schema. Each sample includes a user query, a query category, an explicit reasoning step, a normalized temporal parameter, and a target function. The dataset covers explicit, implicit, behavioral, symptom-based, and metaphorical expressions, which reflect realistic mental health–related user interactions. This resource supports research on intent grounding, temporal reasoning, and reliable function invocation in LLM-based mental health agents and is publicly released to promote reproducibility and future work.
\end{abstract}

\section{Introduction}
LLM–based systems now serve as a core component in many intelligent applications across domains such as education, software engineering, and healthcare~\cite{naveed2025comprehensive}. Recent advances in instruction following and tool use allow LLMs to move beyond conversational output and interact with external systems through structured interfaces~\cite{schick2023toolformer}. This capability has accelerated the development of agentic systems that connect natural language input with real-world data and services~\cite{yao2022react, ouyang2022training, schick2023toolformer}.

Healthcare represents a domain where LLM-based systems offer strong potential but require careful design. Mental health support, in particular, benefits from scalable and personalized assistance while imposing requirements on reliability, interpretability, and controllability. LLM-driven mental health assistants typically operate as non-diagnostic support tools that help users reflect on behavioral patterns and wellbeing indicators rather than providing clinical judgments~\cite{guo2024large, hua2024applying}.

Function calling has emerged as a practical approach for adapting LLMs to domain-specific tasks. Instead of relying on free-form text generation, function calling requires models to map user queries to structured API invocations with explicit arguments~\cite{openai2023functioncalling}. This approach improves reliability, reduces hallucination risk, and enables direct integration with external data sources ~\cite{liu2024toolace}. Despite its adoption in domains such as database querying and software automation, function calling remains underexplored in mental health–oriented applications.

Wearable devices provide continuous access to physiological and behavioral signals such as sleep duration, heart rate, stress indicators, physical activity, and weight trends. These signals offer indirect yet meaningful insight into mental wellbeing and are widely available through consumer devices and standardized health APIs. Effective use of this data requires accurate grounding from natural language intent to data retrieval functions.

At present, no publicly available dataset focuses on function calling grounded in mental health–related wearable sensor data. This paper addresses that gap by introducing a synthetic dataset designed to support research on LLM-based function calling for mental health assistance.

\section{Task Definition}
We study a function-calling task in which an LLM maps a natural language query to a structured API invocation for retrieving wearable sensor data. Each query expresses a user intent to access physiological or behavioral information over a specific temporal range.

Let the input be a natural language query
\[
q \in \mathcal{Q},
\]
where $\mathcal{Q}$ denotes the space of user queries related to wearable health data access.

Let $\mathcal{F} = \{f_1, f_2, \ldots, f_K\}$ be a predefined set of API functions, where each function corresponds to a specific wearable data type (e.g., heart rate, sleep, physical activity). Let $\mathcal{T}$ denote the space of normalized temporal parameters, such as fixed time intervals or relative ranges.

The task is to learn a mapping
\[
\phi : \mathcal{Q} \rightarrow \mathcal{F} \times \mathcal{T},
\]
where the model selects an appropriate function $f \in \mathcal{F}$ and a structured temporal argument $t \in \mathcal{T}$.

In addition, the task may include an intermediate reasoning component
\[
r \in \mathcal{R},
\]
which explains the rationale behind function selection and temporal interpretation. This leads to the extended formulation
\[
\phi(q) = (r, f, t),
\]
where $r$ is not executed by the system but is used to improve interpretability and controllability.

The final output is a structured API call
\[
\texttt{API\_Call}(f, t),
\]
which can be deterministically executed by the downstream system to retrieve the requested wearable sensor data.

This formulation constrains the output space to predefined functions and normalized arguments, emphasizing controlled decision-making and interpretability rather than open-ended text generation.

\section{Motivation: Wearable Sensor Data for Mental Health Assistance}
Wearable sensor data provides longitudinal measurements that reflect daily routines and physiological states. Sleep quality, physical activity, heart rate, stress indicators, and metabolic signals often correlate with mental wellbeing and lifestyle changes. Access to such data enables mental health assistants to support self-reflection and behavior awareness without relying on clinical diagnosis.

Users rarely request wearable data using formal terminology. Instead, they describe symptoms, habits, or feelings that imply a need for specific measurements. A mental health assistant must therefore infer intent from indirect language and map it to relevant data sources. This requirement motivates the need for datasets that capture realistic user expressions tied to wearable sensor retrieval.

\section{Dataset Overview}
This paper introduces \texttt{mind\_call}, a synthetic function-calling dataset designed for wearable sensor data access in mental health–oriented applications. The dataset pairs natural language user queries with structured function calls derived from a standardized health data schema and is publicly available on the Hugging Face platform~\cite{shafimindcall}.

Each sample in \texttt{mind\_call} includes a user query, a query category, an explicit intermediate reasoning step, a normalized temporal parameter, and a target function name. This structure enables fine-grained supervision of intent grounding, temporal interpretation, and function selection.

The \texttt{mind\_call} dataset is designed to support supervised fine-tuning of large language models for function-calling behavior, enabling models to reliably map natural language queries to structured API invocations. In addition, the dataset is well suited for fine-tuning and evaluating LLM-based agent systems that require precise, controllable access to external wearable sensor data as part of multi-step reasoning or decision-making pipelines.

Rather than targeting downstream prediction or clinical diagnosis, \texttt{mind\_call} focuses on controllable intent interpretation and argument normalization. This design makes the dataset particularly suitable for research on trustworthy, interpretable, and deployable agent architectures that integrate LLMs with real-world data sources.

\section{Health Data Schema and Function Definition}
The dataset adopts a standardized wearable health data schema to ensure realism and interoperability. Each function corresponds to a specific physiological or activity-related data type and follows a consistent naming and argument structure. The function set covers sleep, physical activity, cardiovascular signals, metabolic measures, nutrition, stress, and body composition.

The design also supports extensibility. The inclusion of an additional function beyond the current schema illustrates how new sensor modalities can be incorporated without altering the overall task formulation.

\section{Exploratory Data Analysis of the \texttt{mind\_call} Dataset}
To better understand the structure, diversity, and linguistic characteristics of the \texttt{mind\_call} dataset, we conducted an exploratory data analysis (EDA). This analysis examines how user intents are expressed, the semantic and emotional content of queries, and the temporal references that must be resolved for function execution.

The EDA serves two primary purposes. First, it validates that the dataset reflects realistic mental health–oriented interactions rather than synthetic or generic data access requests. Second, it provides insight into the sources of variability and ambiguity that function-calling and agent-based models must handle during fine-tuning and evaluation.

The following subsections analyze query type distributions, domain-specific terminology, affective language, and temporal expression patterns. Together, these analyses highlight the challenges posed by intent grounding and temporal normalization in natural language queries for wearable data access.

\subsection{Query Types and Intent Diversity}
Users express requests for health data in diverse ways that vary in clarity and directness. The \texttt{mind\_call} dataset includes five query categories that reflect common patterns in real-world interactions: explicit, implicit, behavioral, symptom-based, and metaphorical queries. Each category captures a distinct challenge for intent interpretation.

\begin{figure}[t]
    \centering
    \includegraphics[width=\linewidth]{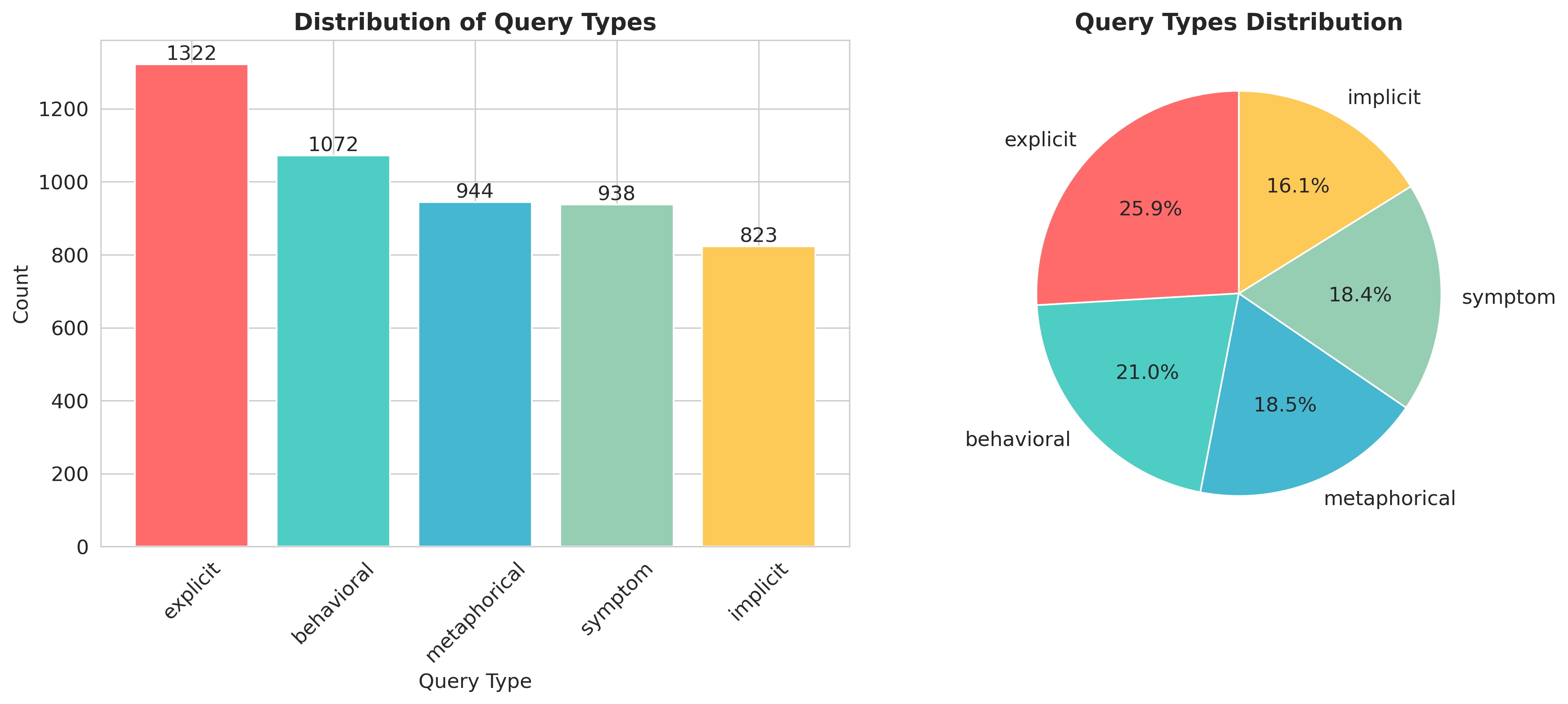}
    \caption{Distribution of query types in the \texttt{mind\_call} dataset. The dataset includes explicit, implicit, behavioral, symptom-based, and metaphorical queries, reflecting diverse user intent expressions.}
    \label{fig:query-type-distribution}
\end{figure}

Figure~\ref{fig:query-type-distribution} shows the distribution of query types across the dataset. The relatively balanced representation ensures coverage of both direct data requests and indirect expressions that require inference, supporting robust model training and evaluation.

This diversity ensures coverage of both direct data requests and indirect expressions that require inference. Balanced representation across categories supports robust model training and evaluation.

\subsubsection{Most Frequent Mental Health--Related Terms}

\begin{figure}[t]
    \centering
    \includegraphics[width=\textwidth]{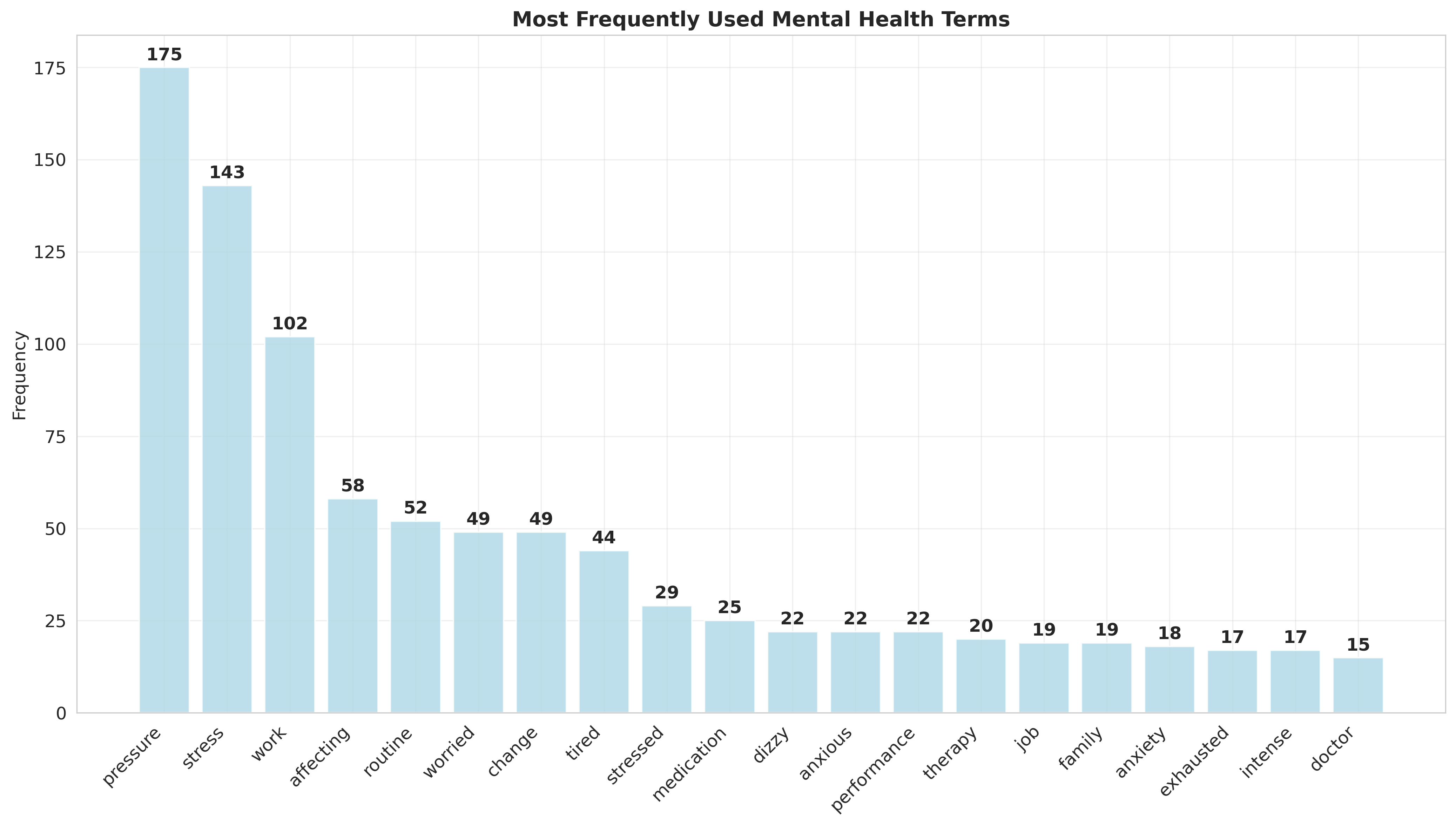}
    \caption[Most frequent mental health--related terms in the dataset]{Most frequent mental health--related terms in the \texttt{mind\_call} dataset, reflecting common topics and concerns aligned with the thesis scope.}
    \label{fig:mental_health_terms}
\end{figure}

To confirm alignment with the mental health support domain, we analyzed the frequency of key terms appearing in user queries. Figure~\ref{fig:mental_health_terms} shows the most frequent mental health--related terms, including sleep, stress, anxiety, and exercise. The prevalence of these terms indicates that the dataset reflects realistic counseling and wellness-oriented interactions rather than generic wearable data retrieval requests.

\subsubsection{Most Common Emotional Terms in Queries}

\begin{figure}[t]
    \centering
    \includegraphics[width=\textwidth]{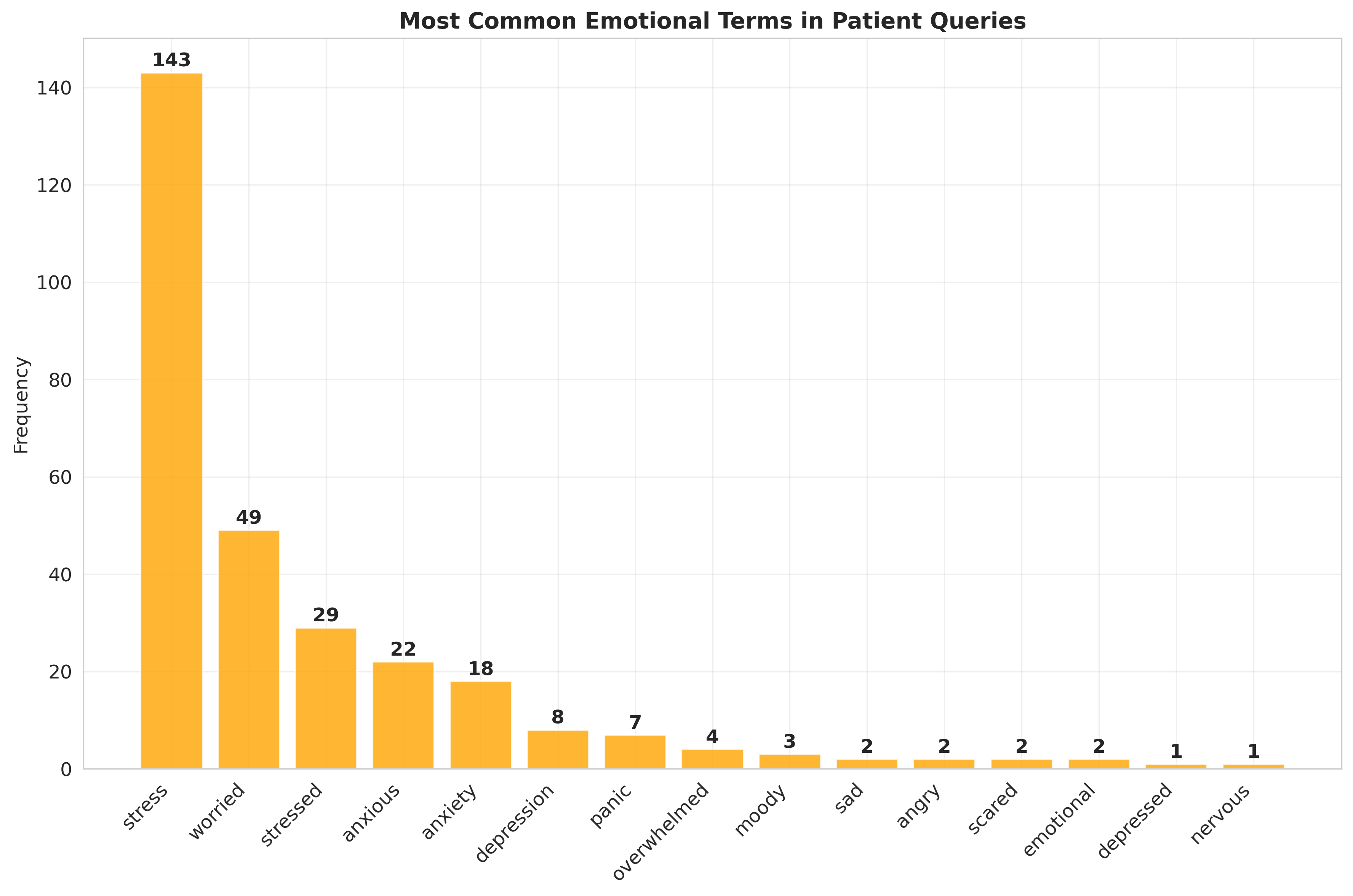}
    \caption[Most common emotional terms in user queries]{Most common emotional terms appearing in user queries, highlighting the affective context of interactions in the \texttt{mind\_call} dataset.}
    \label{fig:emotional_terms}
\end{figure}

Beyond topical keywords, we examined affective vocabulary to characterize the emotional context that often accompanies user requests. Figure~\ref{fig:emotional_terms} presents the most common emotional terms observed in the dataset, with frequent mentions of worry, tiredness, and energy-related states. This distribution highlights the importance of intent interpretation that goes beyond literal keyword matching, particularly in mental health–oriented interactions.

\subsubsection{Temporal Expression Distribution and Mapping}

\begin{figure}[t]
    \centering
    \includegraphics[width=0.95\textwidth]{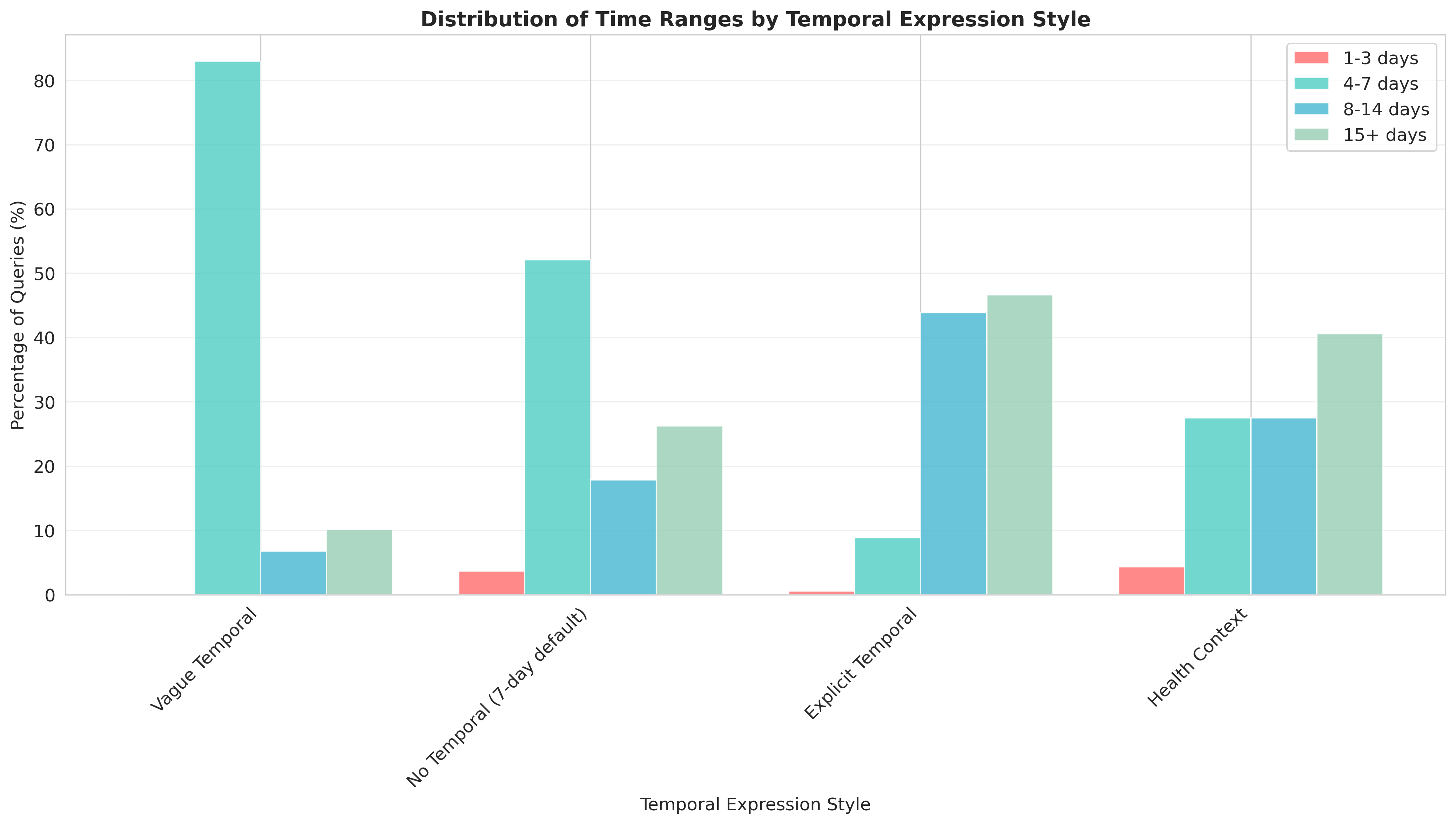}
    \caption[Temporal expression distribution in queries]{Distribution of temporal expressions across explicit, relative, and vague forms. All expressions are normalized to a numeric \texttt{numdays} parameter (default = 7 when unspecified).}
    \label{fig:temporal_expression_distribution}
\end{figure}

Temporal grounding is a core component of the function-calling task, as every API invocation ultimately resolves to a numeric \texttt{numdays} parameter. Figure~\ref{fig:temporal_expression_distribution} categorizes the temporal expressions observed in the dataset, ranging from explicit durations (e.g., ``past 7 days'') to relative references (e.g., ``yesterday'', ``last week'') and vague expressions (e.g., ``recently'').

As described earlier, all temporal expressions are normalized to a numeric \texttt{numdays} value, with a default of 7 days applied when the time span is unspecified. This normalization enables consistent downstream execution while preserving the variability of natural language time references.

\section{Explicit Reasoning Annotation}

\begin{figure}[t]
    \centering
    \includegraphics[width=\textwidth]{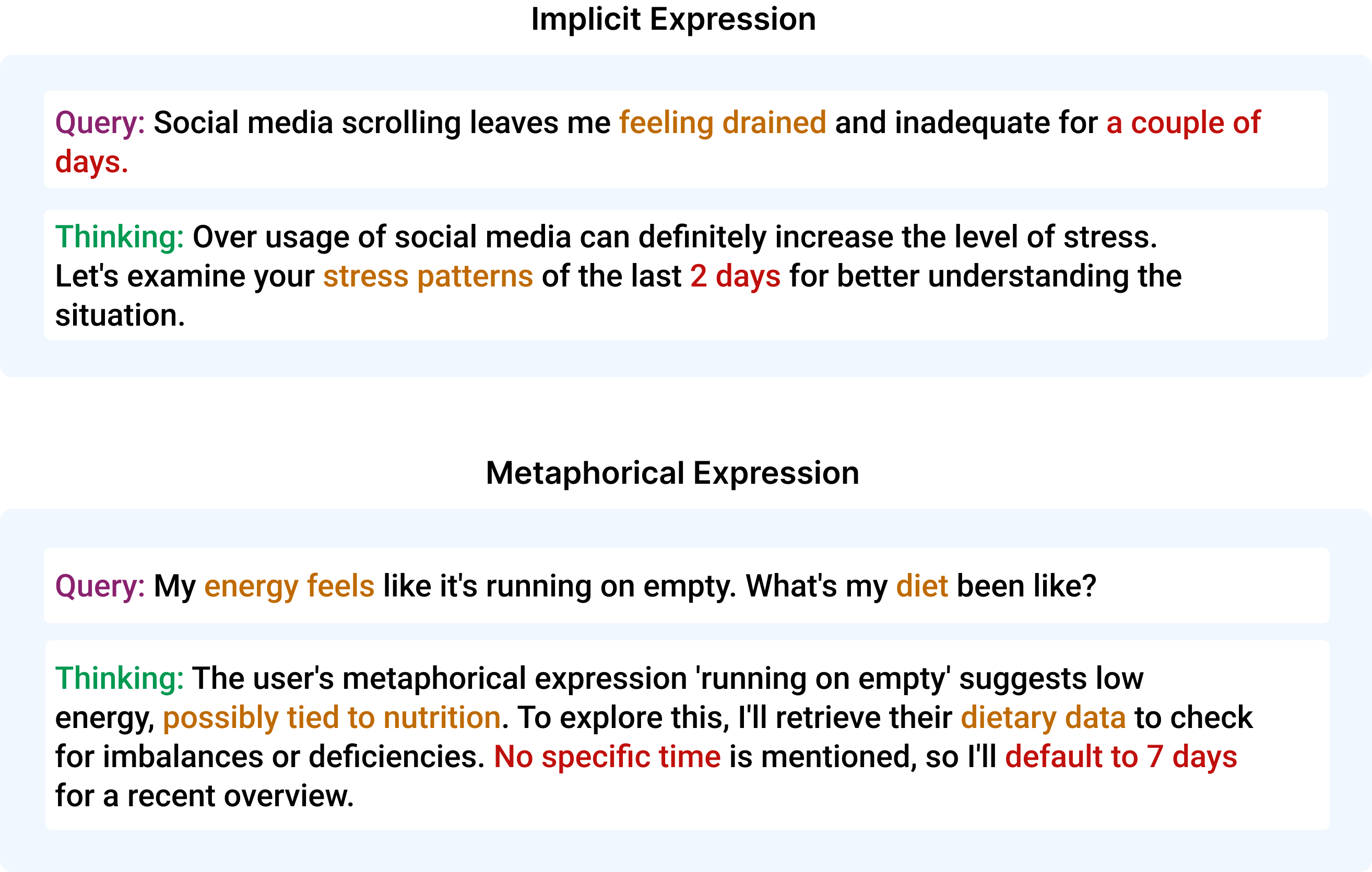}
    \caption[Examples of query--thinking correlation]{Examples illustrating how user queries are paired with corresponding \textit{thinking} annotations in the \texttt{mind\_call} dataset. The reasoning process explains intent interpretation, function selection, and temporal normalization.}
    \label{fig:query_thinking_examples}
\end{figure}

In addition to standard function-calling fields, the \texttt{mind\_call} dataset includes an explicit \textit{thinking} annotation that captures intermediate reasoning prior to function invocation. This reasoning provides a structured explanation of how a user query is mapped to a function and its associated temporal arguments.

Each \textit{thinking} annotation follows a consistent pattern. It first interprets the user’s intent, then justifies the selected function, and finally resolves the temporal constraints expressed in the query. The reasoning is expressed in natural language and is designed to remain concise and task-focused.

The annotation scheme draws inspiration from structured function-calling datasets such as \textit{NousResearch/hermes-function-calling-v1}~\cite{nousresearch_hermes}. While such datasets define clear function schemas, they typically omit an explicit reasoning phase before execution. Recent work on intermediate reasoning and test-time compute~\cite{guo2025deepseek,tian2025think} shows that exposing this reasoning step can improve accuracy, robustness, and interpretability. These findings are consistent with prior work on chain-of-thought prompting~\cite{wei2022chain}, which demonstrates the value of explicit reasoning in structured decision-making tasks.

The \textit{thinking} annotation serves multiple roles within the dataset. It makes the model’s decision process explicit, grounds temporal normalization in the original query, and encourages step-by-step reasoning across diverse query types. These properties support both supervised fine-tuning for function-calling behavior and training of agent-based systems that rely on controlled tool use.

Figure~\ref{fig:query_thinking_examples} illustrates representative examples from the dataset. The examples show how natural language queries are transformed into structured reasoning steps that guide function selection and argument resolution.

\section{Dataset Generation Process}
The dataset was generated through a controlled multi-stage pipeline that uses multiple large language models to produce diverse and realistic samples. Each model received structured prompts that specified the target function, query category, reasoning requirements, and temporal normalization rules. Post-processing removed duplicates and enforced structural consistency.

This process mitigates stylistic bias from individual models and increases linguistic diversity across samples.

\section{Ethical and EDI Considerations}
The dataset consists entirely of synthetic samples and contains no personal identifiers or demographic attributes. It does not support diagnosis or treatment recommendations. All queries and functions focus on data access rather than clinical interpretation.

The dataset is limited to English and does not capture cultural variation in health-related language. Future extensions should address multilingual and cross-cultural representation.

\section{Limitations and Future Works}
The dataset focuses on single-function invocation and wearable sensor data retrieval. It does not model multi-step planning or longitudinal interaction. Future work can extend the dataset to multi-function workflows, additional sensor modalities, and multilingual settings.

\section{Conclusion}
This paper introduced \texttt{mind\_call}, a synthetic function-calling dataset designed to support controlled access to wearable sensor data in mental health–oriented applications. The dataset formalizes the mapping from natural language queries to structured API calls, including explicit annotations for intent, temporal normalization, and intermediate reasoning.

Through exploratory analysis, we demonstrated that \texttt{mind\_call} captures diverse query types, realistic mental health terminology, and varied temporal expressions. These characteristics make the dataset suitable for supervised fine-tuning and evaluation of function-calling models and agent-based systems that require precise and interpretable tool use.

We expect \texttt{mind\_call} to serve as a useful resource for research on LLM-based agents, controllable reasoning, and reliable integration of language models with real-world health data sources.

\bibliographystyle{plain}
\bibliography{references}

\end{document}